\pdfoutput=1
\documentclass[10pt,twocolumn,letterpaper]{article}

\usepackage[accsupp]{axessibility}
\usepackage[pagenumbers]{wacv} 

\usepackage{graphicx}
\usepackage{amsmath}
\usepackage{amssymb}
\usepackage{booktabs}
\usepackage{dutchcal}
\usepackage{booktabs}
\usepackage{multirow}
\usepackage{makecell}
\usepackage{xcolor,colortbl}
\usepackage{graphicx}
\usepackage{multirow}
\usepackage{array}
\usepackage{makecell}

\usepackage[dvipsnames]{xcolor}
\usepackage{bbm}
\usepackage{adjustbox}
\newcommand{\equalcontrib}{\textsuperscript{\tiny 1}}

\usepackage[pagebackref,breaklinks,colorlinks]{hyperref}

\usepackage[capitalize]{cleveref}
\crefname{section}{Sec.}{Secs.}
\Crefname{section}{Section}{Sections}
\Crefname{table}{Table}{Tables}
\crefname{table}{Tab.}{Tabs.}

\def\wacvPaperID{} 
\def\confName{WACV}
\def\confYear{2026}
\DeclareUnicodeCharacter{2032}{\ensuremath{'}}

\begin{document}
\title{Improvise, Adapt, Overcome — Telescopic Adapters for Efficient Fine-tuning of Vision Language Models in Medical Imaging}


\author{
Ujjwal Mishra\equalcontrib \quad
Vinita Shukla\equalcontrib \quad
Praful Hambarde \quad
Amit Shukla \\
\\
Centre for Artificial Intelligence and Robotics, Indian Institute of Technology Mandi, India \\
{\tt\small ujjwalmishra238@gmail.com}, 
{\tt\small \{d23097@students., praful@, amitshukla@\}iitmandi.ac.in} \\
}
\maketitle
\footnotetext[1]{\scriptsize These authors contributed equally.}
\begin{abstract}
Adapting Vision Language Segmentation Models (VLSMs) to medical imaging domains requires significant computational overhead when using conventional fine-tuning approaches. Existing Parameter-Efficient Fine-Tuning (PEFT) methods apply uniform adapter dimensions across all transformer layers, leading to suboptimal parameter allocation and reduced adaptation efficiency. We introduce Telescopic Adapters, a novel PEFT framework that employs depth-aware scaling to progressively increase adapter capacity from shallow to deep transformer layers. Our method integrates lightweight bottleneck modules within CLIPSeg's vision and text encoders, with adapter dimensions dynamically scaled based on layer depth and semantic relevance. Using only 613k trainable parameters—244× fewer than end-to-end fine-tuning, Telescopic Adapters achieve superior performance across five diverse medical datasets spanning polyp segmentation, skin lesion detection, and breast ultrasound imaging. Comprehensive ablation studies demonstrate that deeper layers require substantially more adaptation capacity than shallow layers, validating our telescopic scaling hypothesis. Our approach establishes a new paradigm for efficient medical VLSM fine-tuning, enabling deployment in resource-constrained clinical environments while maintaining competitive segmentation accuracy.
\end{abstract}
\vspace{-2ex}
\section{Introduction}
\label{sec:intro}
Medical image segmentation is a dense prediction task that involves labeling each pixel in an image to identify anatomical structures or areas affected by disease \cite{defibation}. With the increasing use of computer-aided diagnosis (CAD) in the last decade \cite{CAD}, it has become an important tool in diagnosis \cite{diag}, treatment planning \cite{treat3}, and accurate patient monitoring \cite{moni}.
\begin{figure}[h]
\begin{center}
\includegraphics[scale = 0.48]{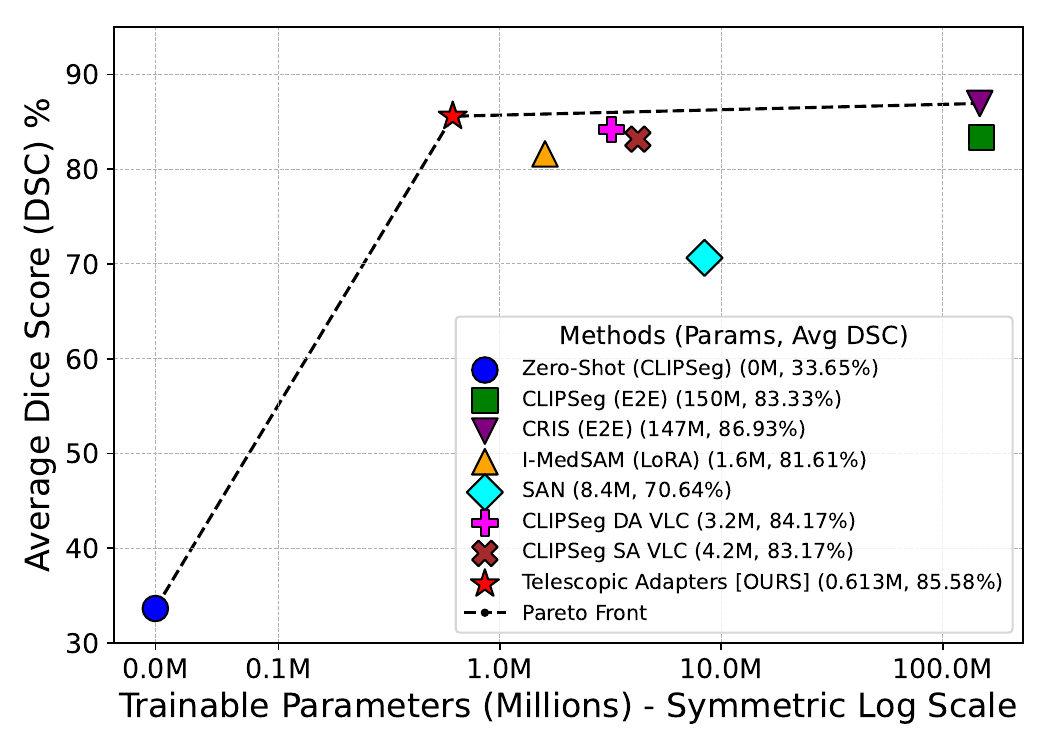}
\end{center}
\vspace{-12pt}
\caption{Performance vs. Parameter Efficiency plot. Average Dice Score (DSC \%) across five datasets is plotted against the number of trainable parameters (in millions) on a symmetric log scale. Our method (Telescopic Adapters) and other parameter-efficient approaches are compared against End-to-End (E2E) fine-tuning and a Zero-Shot baseline. The dashed line indicates the Pareto front, showing the optimal trade-off between performance and parameter cost.}
\label{fig:plott}
\vspace{-12pt}
\end{figure}
The success of deep learning in CAD \cite {suc5, suc6} has translated into substantial improvements in medical image segmentation models across diverse imaging modalities, anatomies, and diseases \cite{good}. However, these models are typically trained on specific foreground classes and imaging modalities, limiting their adaptability to new images or additional information sources \cite{general}. Scaling up data for specific segmentation tasks remains challenging due to high costs of data collection and annotation \cite{data_prob}, posing challenges for real-world clinical deployment where imaging protocols and diagnostic needs vary widely.

Recent efforts have explored incorporating textual information as auxiliary supervision to overcome these limitations \cite{vlsmadap}. Medical images are frequently accompanied by textual reports containing detailed descriptions of anatomical structures, pathological findings, and spatial relations. These text-based cues complement visual information, particularly when annotations are scarce or image quality is suboptimal \cite{vnlm3}.
Vision Language Models (VLMs), such as CLIP \cite{clip}, have emerged as powerful foundations for multimodal approaches. CLIP employs contrastive loss on large-scale image-text pairs to align semantically related embeddings in a shared representation space. Building on these capabilities, VLSMs \cite{vlsm2} extend VLMs to dense prediction tasks. CLIPSeg \cite{clipseg} combines frozen CLIP encoders with lightweight transformer-based decoders for open-vocabulary segmentation, while Yu \textit{et al.} \cite{yu} leverage CLIP with self-supervised mask proposal networks for zero-shot referring image segmentation.
\par Although these models achieve impressive results in open domains \cite{clipseg, cris, open4}, their zero-shot performance on medical image segmentation remains limited. This performance gap necessitates fine-tuning VLSMs on domain-specific medical datasets to effectively capture clinical semantics and imaging characteristics \cite{finetune3}. However, conventional fine-tuning of large-scale VLMs, often comprising hundreds of millions to billions of parameters, is computationally expensive and memory intensive \cite{finetunig}. Such resource demands pose significant challenges for deployment in clinical environments, particularly in low-resource or mobile settings.
Parameter-Efficient Fine-Tuning (PEFT) methods address these limitations by adapting pretrained models through modifying only a small fraction of parameters. Among widely adopted PEFT techniques are Adapters \cite{adaptoer} and Low-Rank Adaptation (LoRA) \cite{lora}. Adapter-based methods insert lightweight, trainable modules within frozen model backbones, enabling task-specific adaptation with minimal computational overhead. Adapter strategies have been explored more extensively than LoRA in vision-language contexts, making them particularly relevant for VLSMs \cite{ca}.
To address these limitations, we introduce \textit{Telescopic Adapters} for fine-tuning VLSMs on medical image segmentation. Our approach augments CLIPSeg with learnable adapter modules and a depth-aware scaling mechanism that progressively increases adapter capacity from shallow to deep layers. Using only 613k trainable parameters—244$\times$ fewer than E2E fine-tuning—our method outperforms existing parameter-efficient approaches as shown in Fig \ref{fig:plott}. The main contributions of this work are:

\begin{itemize}
\item We introduce novel telescopic adapter modules with depth-aware scaling for fine-tuning VLSMs to  domain-specific smaller datasets using minimal learnable parameters.
\item Our experiments and results on medical datasets with diverse modalities indicate that our telescopic adaptation strategy outperforms both end-to-end fine-tuning and existing PEFT methods for VLSMs.
\item We provide comprehensive studies on Adapter dimensions and positioning, demonstrating that progressive scaling and strategic placement result in superior performance with minimal parameter overhead.
\end{itemize}

\section{Related Work}
\subsection{Vision Language Models}
Vision language models (VLMs) have shown notable progress, enabling cross-modal understanding spanning diverse vision tasks \cite{59, 60}. CLIP \cite{clip}, trained on large-scale image–text pairs, has demonstrated strong generalization and zero-shot capabilities \cite{16}. Open source model such as the OpenCLIP implementation \cite{openclip} have further supported large-scale pretraining and reproducibility. Beyond classification, CLIP-based models have been applied to tasks that involve region segmentation and classification through text-driven supervision \cite{18}. These developments have motivated extensions into the medical domain. Models such as PubMedCLIP \cite{14} and MedCLIP \cite{medclip} incorporate biomedical text corpora and improve CLIP alignment in clinical contexts. Other variants \cite{48} introduce self-supervised learning, cross-view consistency, and domain-specific enhancements to improve zero-shot performance in low-resource medical environment. Among these, BiomedCLIP \cite{biomed} demonstrates strong performance in multiorgan retrieval and classification tasks.

Building on the success of CLIP, recent VLSMs aim to take advantage of textual prompts for dense prediction tasks. CLIPSeg \cite{clipseg} enhances CLIP’s architecture by introducing a lightweight transformer decoder that fuses vision and text representations to enable prompt-driven segmentation. Unlike retrieval-based methods, CLIPSeg supports both zero-shot and fine-tuned segmentation scenarios and accommodates ViT backbones. In contrast, Wang \textit{et al.} \cite{cris} formulates segmentation as a region retrieval problem, employing a CNN-based CLIP vision encoder to align visual features with textual queries. While CRIS \cite{cris} has shown promise in localizing relevant regions, its reliance on a convolutional backbone limits scalability and integration with modern transformer-based pipelines. BiomedCLIPS \cite{biomed} further extends this direction by adapting its encoders for segmentation, though it lacks an E2E pretrained decoder and hence is typically used with task-specific fine-tuning.

\subsection{Parameter Efficient Fine Tuning}
The increasing scale of VLMs has improved cross-modal generalization across a wide range of visual tasks. However, these models are predominantly trained on broad open-domain datasets and often underperform when applied to specialized domains such as medical imaging. Domain adaptation strategies address this limitation by transferring knowledge from a source domain to a target domain, typically with limited supervision \cite{82}. Conventional approaches include unsupervised domain adaptation through pseudo-labeling \cite{43}, style transfer \cite{37}, or contrastive and adversarial learning \cite{62}, while more recent work explores test time and online adaptation for dynamic deployment scenarios \cite{611}. Despite their flexibility, these methods frequently incur considerable computational cost and instability during inference.

Given the prohibitive cost of retraining large-scale models in full, PEFT has emerged as an effective alternative for adapting VLMs to downstream tasks with minimal computational burden. PEFT methods reduce training and memory costs by updating only a small subset of parameters, leaving the majority of pretrained model weights frozen \cite{333}. These methods can be broadly classified into four categories: selective tuning, adapters, prompt-based tuning, and low-rank adaptation. Selective tuning methods restrict optimization to specific components within the architecture.  Ben \textit{et al.} \cite{bitfit} fine-tunes only the bias terms of the attention and feedforward layers, while other approaches apply structured pruning to derive task-specific subnetworks \cite{ prune2}.

Adapter-based tuning introduces small trainable modules into the frozen backbone to modulate intermediate representations. These modules are typically lightweight bottleneck architectures inserted between transformer sublayers \cite{adaptoer}. Adapter tuning has shown effectiveness in vision language tasks \cite{ex_adap2}, enabling modular transfer and multitask learning. Variants such as CLIP-Adapter \cite{ca} adapt pretrained features through learned visual pathways that integrate both adapted and original representations. However, insertion of an adapter increases the number of model parameter inferences, which may lead to higher latency. Recent efforts aim to address this drawback through compression and adapter fusion techniques \cite{fusion}.

Prompt tuning replaces manually designed prompts with learnable embeddings. Initially applied to language models, this approach has been extended to vision-language settings by optimizing prompts or visual tokens \cite{prompt1}. These tunable embeddings, when prepended to inputs or injected into intermediate layers, steer the model toward domain-relevant behavior with minimal parameter updates. Prompt tuning is particularly attractive for few-shot and zero-shot settings \cite{fs1}.

Low-Rank Adaptation (LoRA) \cite{lora} represents another class of PEFT, where trainable low-rank matrices are injected into linear projections of the frozen model to capture task-specific weight updates. By design, LoRA allows for merging the adapted parameters with the original weights before inference, resulting in no additional latency or memory usage. LoRA assumes task-specific modifications reside in a low-dimensional subspace, enabling efficient fine-tuning. LoRA extension includes rank-adaptive variants \cite{lora2}, improved optimization strategies \cite{perform1, perform2}, and quantized versions to reduce memory consumption. Wei \textit{et} al. \cite{imedsam} introduces LoRA to medical image segmentation paradigm for continual learning.  Despite its efficiency, LoRA's performance can be sensitive to the choice of rank and may underperform on tasks requiring more expressive adaptation, especially in high-variance domains like medical imaging.
\section{The Proposed Method}
\begin{figure*}[h]
\begin{center}
\includegraphics[scale = 0.175]{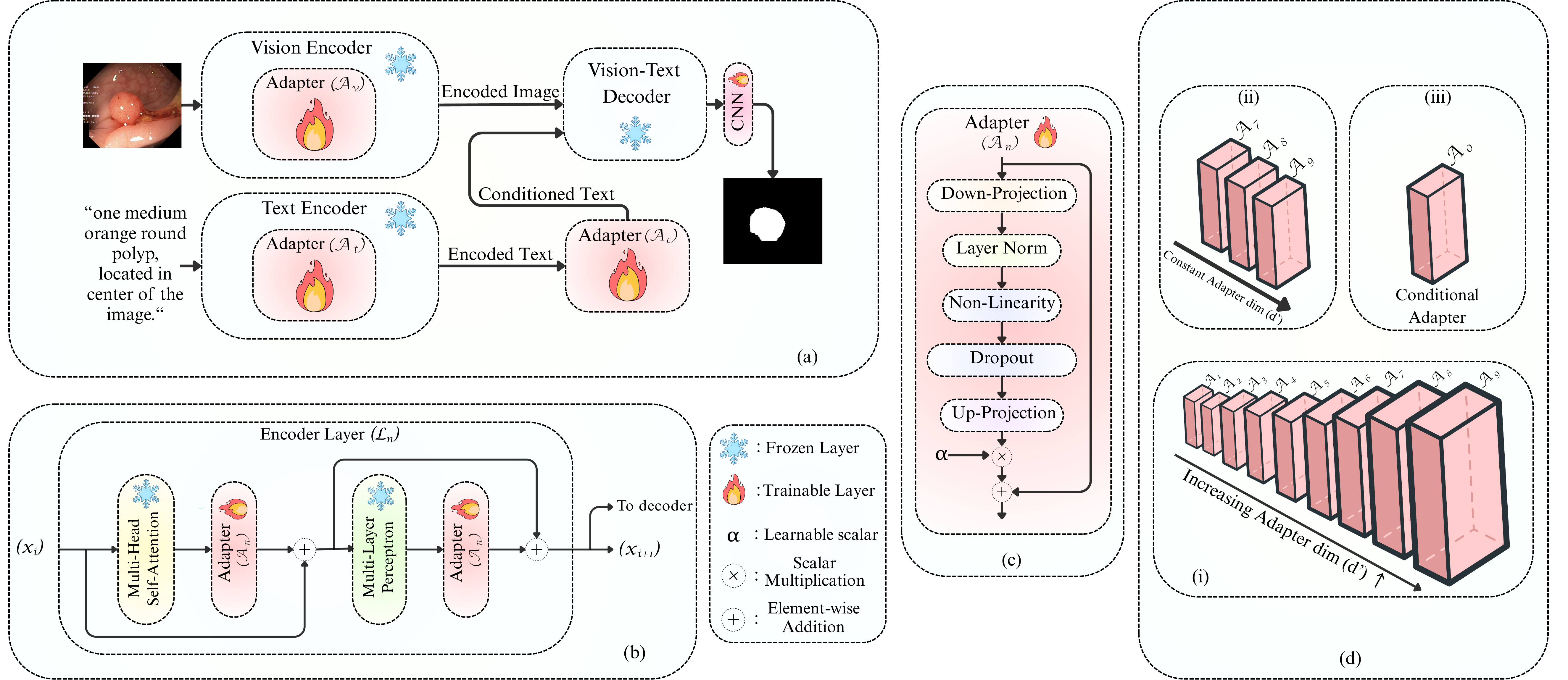}
\end{center}
\vspace{-12pt}

\caption{Overview of the proposed telescopic adaptation framework. 
(a) Overall architecture of the modified CLIPSeg model with integrated adapter modules. 
(b) Placement of adapters within the encoder branch, inserted after the multi-head self-attention and MLP sublayers. 
(c) Proposed adapter formulation, where the learnable scalar $\alpha$ regulates the contribution of the adapter at each layer. 
(d) Telescopic Adapters across different modalities: (i) Vision branch, (ii) Text branch, and (iii) Conditional branch, illustrating progressive dimension allocation and selective adaptation.}
\label{fig:arch}
\vspace{-12pt}
\end{figure*} 
As shown in Fig.~\ref{fig:arch} (a), we extend the CLIPSeg architecture by introducing lightweight adapter modules within the transformer encoders of both the vision and text branches, as well as a conditional adapter after the text projection. 
Adapters are selectively inserted into transformer layers at positions preceding residual addition, allowing localized modulation of representations at both attention and feedforward sublayers as shown in Fig.\ref{fig:arch} (b). Each adapter is designed as a bottleneck projection with a learnable residual scaling parameter to control its influence on the layer output.
To improve adaptation efficiency, we introduce \textit{telescopic adaptation} strategy to dynamically adjust the dimension of the adapter between layers based on their depth and relevance to the task. Deeper layers, which encode higher-level semantic representations, are assigned progressively larger bottleneck dimensions, while early layers receive minimal capacity to reduce redundancy (see Fig.\ref{fig:arch} (d)).
Finally, we append a lightweight CNN layer with 2 layers and only 48 trainable parameters after the CLIPSeg decoder. This module enhances spatial refinement of the segmentation mask without introducing significant computational overhead.

\subsection{Adapter Placement}
\label{sec:adapter_placement}
We integrate lightweight adapter modules selectively across the vision, text, and conditional pathways to enable efficient and localized task-specific adaptation. These adapters are inserted within the transformer encoder blocks of the model without modifying the pretrained backbone weights as shown in Fig.\ref{fig:arch} (a).

Let $\mathcal{A}_v$, $\mathcal{A}_t$, and $\mathcal{A}_c$ denote the adapters in the vision, text, and conditional branches, respectively. Each adapter $\mathcal{A}_n$ is inserted in the $n^{th}$ transformer encoder layer $\mathcal{L}_n$ of its corresponding modality. Adapters are applied before each residual addition point in $\mathcal{L}_n$, resulting in two adapter applications per encoder layer: once after the multi-head self-attention sublayer and once after the feed-forward MLP sublayer  as shown in Fig.\ref{fig:arch} (b). This placement ensures that task-specific modulation occurs before residual addition \cite{adaptoer}.
\\
\textbf{Vision Adapters} ($\mathcal{A}_v$) are inserted into the first nine transformer layers $\{ \mathcal{L}_1, \mathcal{L}_2, \ldots, \mathcal{L}_9 \}$. CLIPSeg utilizes only the intermediate representations of these layers within its decoder, and as such, adapting these specific layers allows the model to progressively adjust the visual features passed to the decoder. Each layer is augmented with two adapter modules placed before the two residual connections, resulting in 18 adapter instances on the vision encoder. This configuration ensures that each encoder block refines its activation before passing it both to the decoder and to the next layer in the stack \cite{clipseg}.

\textbf{Text Adapters} ($\mathcal{A}_t$) are inserted into the final three layers: $\mathcal{L}_7$, $\mathcal{L}_8$ and $\mathcal{L}_9$. These layers capture the most semantically rich and task-relevant features, making them ideal for lightweight adaptation. Earlier layers, which primarily encode syntactic and general-purpose features, are left unaltered to avoid unnecessary interference. As with the vision encoder, each selected text layer contains two adapters, one after multi-head self-attention sublayer and one after the feed-forward MLP sublayer, totaling six $\mathcal{A}_t$ modules in the text pathway.

\textbf{Conditional Adapter} ($\mathcal{A}_c$) is applied once to the output of the text encoder after projection. This adapter enables minimal but targeted transformation of the final joint text representation before it is passed to the decoder, supporting a better alignment between the textual context and the segmentation objective.

This selective and structured placement leverages the encoder-decoder architecture of CLIPSeg, ensuring that only the most relevant layers are adapted to suit the downstream dense prediction task.

\subsection{Adapter Formulation}
Building on the formulation introduced in \cite{adaptoer}, we design an adapter module that introduces non-linearity, dimensionality compression, and task-specific flexibility into the network without modifying the pretrained backbone as presented Fig.\ref{fig:arch} (c). The adapted representation is computed as:
\begin{equation}
\mathbf{f}' = \mathbf{f} + \alpha \cdot \mathbf{W}_\text{up} \left( \text{Dropout} \left( \sigma \left( \text{Norm} \left( \mathbf{W}_\text{down} \cdot \mathbf{f} \right) \right) \right) \right)
\end{equation}

Here, \( \mathbf{f} \in \mathbb{R}^{\dots \times d} \) represents the input feature from the pretrained model, and \( \mathbf{f}' \in \mathbb{R}^{\dots \times d} \) is the adapted output. The transformation is defined by a bottleneck structure, where \( \mathbf{W}_\text{down} \in \mathbb{R}^{d \times d'} \) reduces the dimensionality and \( \mathbf{W}_\text{up} \in \mathbb{R}^{d' \times d} \) projects it back to the original space. The bottleneck dimension ($d'$) is defined as the following inside the adapter:

\begin{equation}
d' = \max(8, \min(d_{\text{adapter}}, \frac{d}{4}))
\label{adapter_class}
\end{equation}
We compute \( d_{\text{adapter}} \) on a per-layer basis to maintain efficiency and expressiveness. A detailed discussion on this adaptive choice is presented in the next section.
The compressed features are then normalized along the feature dimension to ensure stable gradient propagation, followed by a \textit{SiLU} activation function, which introduces smooth, input-aware non-linearity particularly suited to compact transformation layers \cite{SILU2}. Then a fixed rate \textit{dropout} (p = 0.1) is applied to regularize the transformation and reduce overfitting.

We also introduce a learnable scalar parameter \( \alpha \in \mathbb{R} \), initialized to 0.1. This small initial value is crucial for training stability, as it ensures the untrained adapter's random weights do not destabilize the pretrained model's representations at the start of fine-tuning. This residual scaling allows the model to softly regulate the extent of task-specific adaptation at each layer. Importantly, it also improves the interpretability of the adaptation mechanism, as higher values of \( \alpha \) reflect layers where the adapter contributes more prominently.

\subsection{Telescopic Adaptation}
\label{sec:telescopic_adaptation}
To improve parameter efficiency and adaptivity, we introduce a telescopic adaptation mechanism that modulates the adapter size across transformer layers based on their depth and functional importance. Rather than employing uniform adapter widths throughout the network, this strategy assigns smaller bottleneck dimensions to early layers and progressively increases the capacity toward deeper layers.

Transformer encoders follow a coarse-to-fine representational hierarchy. The early layers predominantly capture low-level or syntactic features, while the deeper layers encode more abstract task-specific semantics \cite{dim}. Imposing large adapters on all layers uniformly not only adds redundancy but may also destabilize training. We address this by dynamically computing the per-layer adapter dimension \( d_{\text{adapter}} \) from a base value \( d_{\text{base}} = 64 \), which we empirically select to balance expressiveness and stability. Larger values were found to cause gradient explosion in the early stages of training.

\textbf{Vision Adapters.} Let \( \mathcal{L}_v \) denote the number of visual encoder layers with $\mathcal{A}_v$. For the $i^{th}$ layer ($1 \leq i \leq \mathcal{L}_v$), we scale the adapter dimension progressively with depth using a layer-dependent factor:
\begin{equation}
d_{\text{adapter}}^{(v,i)} = \max\left(8,\left\lfloor \frac{1}{2} \cdot d_{\text{base}} \cdot \frac{i}{\mathcal{L}_v} \right\rfloor \right)
\label{telescopic_vision}
\end{equation}
This formulation ensures a minimal adapter width of 8 and gradually increases the dimensionality to support richer adaptation in deeper layers where features are more task-relevant.

\textbf{Text Adapters.} In contrast to vision, we restrict adaptation in the text encoder to its final three layers, which carry semantically rich embeddings crucial to conditioning. Here, we opt for a statically reduced adapter dimension across all three layers:
\begin{equation}
d_{\text{adapter}}^{(t)} = \max\left(8,\left\lfloor \frac{d_{\text{base}}}{4} \right\rfloor\right)
\end{equation}
This design imposes a smaller adapter dimension of 16, reflecting the relative stability and abstraction of late-stage language representations without incurring unnecessary overhead.

\textbf{Conditional Adapter.} The conditional adapter, applied after the text projection stage, is configured with an even smaller dimension:
\begin{equation}
d_{\text{adapter}}^{(c)} = \max\left(16,\left\lfloor \frac{d_{\text{base}}}{8} \right\rfloor\right)
\end{equation}
This conservative allocation minimizes interference while allowing for controlled, task-aware modulation of the final joint text embedding before it is passed to the decoder.

 The resulting adapter dimensions are then passed to the formulation in Eq.~(\ref{adapter_class}), where they are further bounded by the input size \( d \) to ensure stability.

\section{Experiments}
\subsection{Datasets and Evaluation Metrics}
To evaluate the effectiveness of our proposed approach, we utilize a diverse collection of medical imaging datasets spanning multiple domains, including both radiology and non-radiology modalities. Following Poudel \textit{et al.} \cite{vlsm3}, we adopt their predefined dataset splits and corresponding text prompts to ensure consistency and comparability in performance benchmarking. Their method generates multiple text prompts per image-mask pair; in our implementation, we randomly select one prompt per sample to construct image-mask-text triplets for evaluation across datasets.

For non-radiology tasks, we utilized three polyp segmentation datasets from endoscopic images: $Kvasir-SEG$ \cite{kvasir}, $ClinicDB$ \cite{clinicbd}, and $BKAI$ \cite{bakai}. We also included the skin lesion segmentation ($ISIC-16$ \cite{isic16}) dataset for skin cancer detection. For radiology, the breast ultrasound image segmentation dataset ($BUSI$~\cite{busi}) was used.

We evaluated the overall performance of the methods using the Dice Similarity Coefficient (DSC \%) and Intersection over Union (IoU \%), with values averaged across each dataset for the three random seed values. All metrics were computed using the MONAI framework \footnote{https://monai.io}.


\subsection{Implementation Details}
The training and inference processes for both the baseline and proposed methods were carried out on an NVIDIA RTX A4000 GPU. Mixed-precision training with floating-point 16 was used, employing a batch size of 32. The initial learning rate for the adapter network is set to $1 \times 10^{-3}$. A scheduler reduced the learning rate by a factor of 0.3 if no decrease in validation loss was observed over 5 consecutive epochs. Training was stopped early if the validation DSC (\%) did not improve in 20 consecutive epochs. All models were optimized using AdamW~\cite{adam} with a weight decay of $1 \times 10^{-3}$. Each experiment was repeated with three different random seed values to assess consistency and account for variability in prompt sampling. The loss function combined Dice and binary cross-entropy losses as shown in Eq.~(\ref{eq:loss}) where scalar  values $\lambda_d $ and $\lambda_{BCE}$ are set to be 1.5 and 1. 
\begin{equation}
L = \lambda_d \cdot L_{\text{Dice}} + \lambda_{BCE} \cdot L_{\text{BCE}}
\label{eq:loss}
\end{equation}

\subsection{Telescopic Adapter Analysis}
To validate the effectiveness of our proposed telescopic adaptation strategy, we perform a comprehensive analysis of the learned adapter parameters after convergence of training. This analysis examines the evolution of the scale parameters \(\alpha\) in the vision and text branches and their distribution across layers to empirically verify the hypothesis that deeper transformer layers require more substantial task-specific adaptation. Fig.~\ref{fig:arch} (d) illustrates the telescopic design: the vision branch (i) uses progressively larger adapter dimensions across layers, the text branch (ii) applies constant dimensions in higher layers, and the conditional branch (iii) employs a single compact adapter after text projection.

\begin{table}[h!]
\centering

\resizebox{\columnwidth}{!}{%
\begin{tabular}{cccc}
\toprule
\textbf{Layer} & \textbf{Adapter Dim} & \textbf{Attention Scale (\(\alpha^V_{A}\))} & \textbf{MLP Scale (\(\alpha^V_{MLP}\))} \\
\midrule
1 & 8  & 0.0465 & 0.0725 \\
2 & 8  & 0.0532 & 0.0905 \\
3 & 10 & 0.0788 & 0.1067 \\
4 & 14 & 0.0880 & 0.1145 \\
5 & 17 & 0.0778 & 0.0956 \\
6 & 21 & 0.0866 & 0.0982 \\
7 & 24 & 0.1098 & 0.1328 \\
8 & 28 & 0.0764 & 0.1164 \\
9 & 32 & 0.0949 & 0.1150 \\
\bottomrule
\end{tabular}%
}
\caption{Layer-wise adapter configurations and learned scale parameters in the vision branch, 
showing the assigned adapter dimensionality and the final Attention (\(\alpha^V_{A}\)) 
and MLP (\(\alpha^V_{MLP}\)) scaling factors after training convergence.}
\label{tab:vision_adapters}
\end{table}
\vspace{-0.40cm}
\begin{table}[h!]
\centering

\resizebox{\columnwidth}{!}{%
\begin{tabular}{cccc}
\toprule
\textbf{Layer} & \textbf{Adapter Dim} & \textbf{Attention Scale (\(\alpha^T_{A}\))} & \textbf{MLP Scale (\(\alpha^T_{MLP}\))} \\
\midrule
7 & 16 & 0.0816 & 0.1061 \\
8 & 16 & 0.1063 & 0.0924 \\
9 & 16 & 0.0962 & 0.1070 \\
\bottomrule
\end{tabular}%
}
\caption{Layer-wise adapter configurations and learned scale parameters in the text branch, 
showing the assigned adapter dimensionality and the final attention (\(\alpha^T_{A}\)) 
and MLP (\(\alpha^T_{MLP}\)) scaling factors after training convergence.}
\label{tab:text_adapters}
\end{table}

\subsubsection{Scale Parameter Evolution and Regularization}
The learned scale parameters were effectively regularized during training. In the vision branch (Table~\ref{tab:vision_adapters}), \(\alpha^V_{A}\) and \(\alpha^V_{MLP}\) range from 0.046 to 0.133, while in the text branch (Table~\ref{tab:text_adapters}), \(\alpha^T_{A}\) and \(\alpha^T_{MLP}\) lie between 0.082 and 0.107. In the vision branch, 12 out of 18 values converged below the initialization threshold of 0.1, while six exceeded it, indicating targeted, high-impact adaptation in specific layers. In the text branch, 3 out of 6 values fell below this threshold, while three surpassed it. On average, the vision parameters show a reduction of 0.81\% from their initial value, whereas the text parameters show a smaller mean reduction of 0.17 \%. This controlled convergence pattern suggests that the scaling mechanism successfully modulated adaptation and maintained stability, confirming the robustness of our residual scaling formulation in Eq.(~\ref{adapter_class}).
\subsubsection{Layer-wise Contribution Patterns}
The vision branch exhibits distinct patterns consistent with the telescopic adaptation hypothesis. Early layers (1–3) have lower scales, with \(\alpha^V_{A}\) ranging from 0.046 to 0.079 and \(\alpha^V_{MLP}\) from 0.072 to 0.107, confirming that shallow layers require minimal task-specific modification. Deeper layers show elevated scales, with $\mathcal{L_7}$ reaching the maximum values of \(\alpha^V_{A} = 0.110\) and \(\alpha^V_{MLP} = 0.133\). This indicates that deeper layers capture more task-relevant information.

\subsubsection{Telescopic Dimension Effectiveness}
The correlation between adapter dimension allocation and learned scales supports the telescopic design.$\mathcal{L_7}$, with an assigned dimension of 24 (Eq.~\ref{telescopic_vision}), shows the highest adaptation in both \(\alpha^V_{A}\) and \(\alpha^V_{MLP}\). Early layers with small dimensions (e.g., $\mathcal{L_1}, \mathcal{L_2}$) correspondingly show low scales, confirming that conservative allocation prevents unnecessary parameter usage in shallow layers.

\subsubsection{Attention versus MLP Adaptation}
In the vision branch, \(\alpha^V_{MLP}\) consistently exceeds \(\alpha^V_{A}\), with \(\alpha^V_{MLP}\) ratio  \(\alpha^V_{A}\) ranging from 1.13 ($\mathcal{L_6}$) to 1.70 ($\mathcal{L_2}$), averaging 1.36. This suggests that MLP sublayers are more critical for task-specific adaptation than attention sublayers in dense prediction tasks. In the text branch, adaptation is limited to the last three layers, but unlike the vision branch, the relationship between \(\alpha^T_{A}\) and \(\alpha^T_{MLP}\) is inconsistent. For example, \(\alpha^T_{MLP}\) dominates in $\mathcal{L_7}$ and $\mathcal{L_9}$, while \(\alpha^T_{A}\) is higher in $\mathcal{L_8}$ (\(\alpha^T_{A} = 0.106\) vs. \(\alpha^T_{MLP} = 0.092\)). This suggests that in language processing, the relative contribution of attention versus MLP adaptation varies depending on the layer’s role.

\subsubsection{Parameter Efficiency Validation}
Across both modalities, scale value ($\alpha$) for 15 out of 24 (12 in vision, 3 in text) adapters converged to values below the initialization threshold ($\alpha = 0.10$), demonstrating that the approach avoided over-parameterization. The nine values exceeding the threshold (6 in vision, 3 in text) represent meaningful adaptations that correlate with improved performance, indicating the model emphasized the most task-critical representations.

Overall, this analysis confirms that telescopic adaptation selectively emphasizes semantically critical layers while maintaining parameter efficiency through conservative scaling in less critical regions, validating the coarse-to-fine representational hierarchy underlying our approach.

\subsection{Ablation Studies}
\subsubsection{Ablation Study Over Adapter Placement}

\begin{table}[h!]
\centering
\adjustbox{scale=0.86}{
\begin{tabular}{llccc}
\toprule
\multirow{2}{*}{\textbf{Datasets}} & \multirow{2}{*}{\textbf{Metric}} & \multicolumn{3}{c}{\textbf{Adapter Configurations}} \\
\cmidrule(lr){3-5}
& & \textbullet~[498k] & \textbullet\textbullet~[593k] & \textbullet\textbullet\textbullet~[613k] \\
\midrule
\multirow{2}{*}{Kvasir-SEG \cite{kvasir}} & DSC (\%) & 87.35 & 89.67 & \textbf{89.79} \\
                            & IoU (\%) & 80.32 & \textbf{83.62} & 83.50 \\
\midrule
\multirow{2}{*}{BKAI \cite{bakai}}       & DSC (\%) & 85.53 & 87.09 & \textbf{88.38} \\
                            & IoU (\%) & 77.77 & 80.00 & \textbf{81.63} \\
\midrule
\multirow{2}{*}{ClinicDB \cite{clinicbd}}   & DSC (\%) & 85.39 & 91.28 & \textbf{91.67} \\
                            & IoU (\%) & 78.45 & 84.85 & \textbf{85.19} \\
\midrule
\multirow{2}{*}{ISIC-16 \cite{isic16}}    & DSC (\%) & 91.61 & \textbf{92.24} & 92.18 \\
                            & IoU (\%) & 85.30 & \textbf{86.16} & 86.12 \\
\midrule
\multirow{2}{*}{BUSI \cite{busi}}       & DSC (\%) & \textbf{70.35} & 64.33 & 65.90 \\
                            & IoU (\%) & \textbf{62.45} & 57.26 & 59.10 \\
\bottomrule
\end{tabular}
}
\caption{Ablation study on adapter placement across five datasets. The configurations shown are: Vision only branches (\textbullet), Vision and Text branches (\textbullet\textbullet), and Vision, Text, and Conditional branches(\textbullet\textbullet\textbullet). Values in brackets correspond to the total number of trainable parameters per configuration. Best results for each metric are in \textbf{bold}.}
\label{tab:ablation_placement}
\end{table}
\begin{figure*}[h]
\begin{center}
\includegraphics[scale = 0.127]{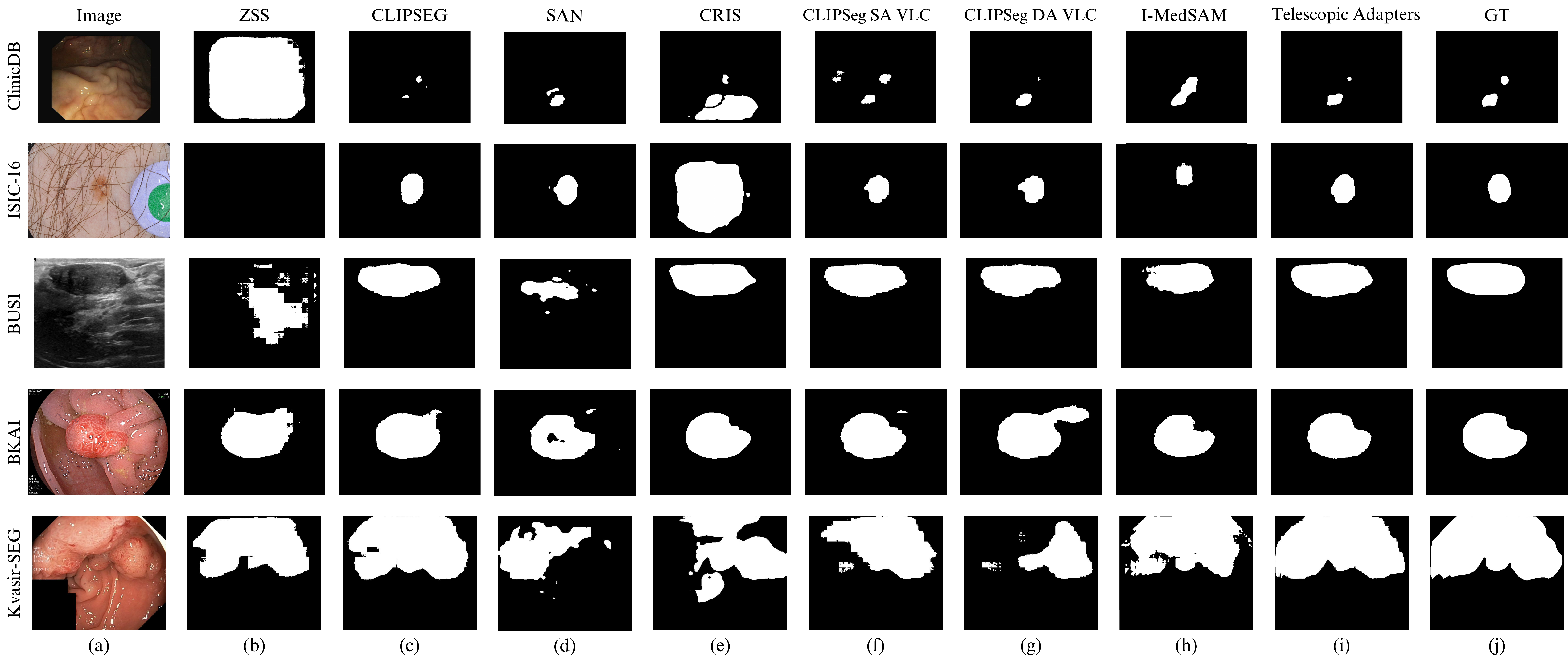}
\end{center}
\vspace{-17pt}

\caption{Segmentation results on samples from ClinicDB \cite{clinicbd}, ISIC-16 \cite{isic16}, BUSI \cite{busi}, BKAI \cite{bakai}, and Kvasir-SEG \cite{kvasir} datasets (rows 1-5 respectively). Column (a) shows the original medical images, followed by segmentation masks from: (b) ZSS (zero-shot segmentation using CLIPSeg \cite{clipseg}), (c) CLIPSeg \cite{clipseg} (E2E fine-tuned), (d) SAN \cite{san}, (e) CRIS \cite{cris} (E2E fine-tuned), (f) CLIPSeg SA VLC \cite{vlsmadap}, (g) CLIPSeg DA VLC \cite{vlsmadap}, (h) I-MedSAM \cite{imedsam}, (i) Telescopic Adapters, and (j) GT (ground truth).}
\label{fig:COMP}

\end{figure*} 
To validate the contribution of each component in our architecture, we conduct an ablation study on adapter placement. The results in Table~\ref{tab:ablation_placement} assess the performance of three models: 1) With adapters in the vision branch only (\textbullet), 2) With adapters in the vision and text branches (\textbullet\textbullet), and 3) our full model incorporating adapters in the vision, text, and conditional branches (\textbullet\textbullet\textbullet).

The addition of text adapters (\textbullet\textbullet) to the vision branch baseline (\textbullet) provides a substantial performance uplift across most datasets, underscoring the importance of textual context. For example, on the ClinicDB \cite{clinicbd} dataset, this addition improves the DSC from 85.39\% to 91.28\% and the IoU from 78.45\% to 84.85\%.

Our proposed model (\textbullet\textbullet\textbullet), which includes the conditional branch, consistently delivers the best or most competitive performance. This final component adds further refinement for a minimal parameter overhead of only 20k, achieving the highest scores for both metrics on the BKAI \cite{bakai} and ClinicDB datasets. Which confirms that the conditional adapters effectively harmonize the features from the vision and text branches.

Notably, for the BUSI \cite{busi} dataset, the vision branch baseline (\textbullet) yields the best results. This suggests that for certain datasets with less complex inter-modal dependencies, the additional branches may not provide a benefit. However, the superior performance of the three (\textit{i.e.} Vision, Text, and Conditional) branches model on approximately four out of five diverse datasets validates it as the most robust and effective configuration, offering an optimal balance between parameter efficiency and high generalization capability.

\begin{table*}[htbp]
\centering
\resizebox{\textwidth}{!}{%
\begin{tabular}{@{}llcccccccc@{}}
\toprule
\multirow{2}{*}{\textbf{Dataset}} & \multirow{2}{*}{\textbf{Metric}} & \multicolumn{1}{c}{\textbf{Zero-Shot}} & \multicolumn{2}{c}{\textbf{End-to-End fine-tuning}} & \textbf{LoRA Fine-Tuning} & \multicolumn{4}{c}{\textbf{Adapter Fine-Tuning}}  \\
\cmidrule(lr){3-3} \cmidrule(lr){4-5} \cmidrule(lr){6-6} \cmidrule(lr){7-10} 
& & \makecell{CLIPSeg \cite{clipseg}\\150M} & \makecell{CLIPSeg \cite{clipseg}\\150M} & \makecell{CRIS \cite{cris}\\147M} & \makecell{I-MedSAM \cite{imedsam}\\1.6M} & \makecell{SAN \cite{san}\\8.4M} & \makecell{CLIPSeg DA VLC \cite{vlsmadap}\\3.2M } & \makecell{CLIPSeg SA VLC \cite{vlsmadap}\\4.2M } & \makecell{Telescopic Adapters \\ \textbf{[OURS]} 613k} \\
\midrule
\rowcolor{gray!10}
\textbf{Kvasir-SEG \cite{kvasir}} & DSC (\%) & 53.42 & 87.69 & 89.43 & 86.36 & 69.58 & 89.06 & 86.80 & \textbf{89.79} \\
\rowcolor{gray!10}
 & IoU (\%) & 42.48 & 81.72 & 83.37 & 78.66 & 58.05 & 82.28 & 79.33 & \textbf{83.50} \\
\textbf{BKAI \cite{bakai}} & DSC (\%) & 30.08 & 85.59 & 92.62 & 88.05 & 66.26 & 87.17 & 83.95 & \textbf{88.38} \\
 & IoU (\%) & 23.96 & 77.52 & 88.30 & 80.87 & 54.58 & 79.74 & 75.61 & \textbf{81.63} \\
\rowcolor{gray!10}
\textbf{ClinicDB \cite{clinicbd}} & DSC (\%) & 36.58 & 88.58 & 93.63 & 88.12 & 81.36 & 89.63 & 89.04 & \textbf{91.67} \\
\rowcolor{gray!10}
 & IoU (\%) & 27.54 & 81.51 & 88.74 & 79.91 & 72.61 & 82.89 & 81.95 & \textbf{85.19} \\
\textbf{ISIC-16 \cite{isic16}} & DSC (\%) & 25.15 & 91.88 & 91.49 & 84.00 & 90.39 & 92.02 & 91.55 & \textbf{92.18} \\
 & IoU (\%) & 20.85 & 85.76 & 85.41 & 75.10 & 83.61 & 85.90 & 85.22 & \textbf{86.12 }\\
\rowcolor{gray!10}
\textbf{BUSI \cite{busi}} & DSC (\%) & 23.01 & 62.91 & 67.50 & 61.52 & 45.61 & 62.95 & 64.52 & \textbf{65.90} \\
\rowcolor{gray!10}
 & IoU (\%) & 20.32 & 55.52 & 60.90 & 56.47 & 35.27 & 54.95 & 56.88 & \textbf{59.10} \\
\bottomrule
\end{tabular}%
}
\caption{
Quantitative benchmark of segmentation performance across five medical datasets: Kvasir-SEG \cite{kvasir}, BKAI \cite{bakai}, ClinicDB \cite{clinicbd}, ISIC-16 \cite{isic16}, and BUSI \cite{busi}. We compare our Telescopic Adapters (OURS) against: (i) the Zero-Shot (ZS) baseline (CLIPSeg \cite{clipseg}), (ii) End-to-End (E2E) fine-tuning (CLIPSeg \cite{clipseg}, CRIS \cite{cris}), (iii) LoRA-based tuning (I-MedSAM \cite{imedsam}), and (iv) other adapter-based methods (SAN \cite{san}, CLIPSeg DA/SA VLC \cite{vlsmadap}). Performance is reported in DSC (\%) and IoU (\%). Parameter counts shown for PEFT methods and End-to-End fine-tuning are trainable; for   Zero-Shot segmentation, they are the total model parameters.
}
\label{tab:medical_segmentation_results}
\end{table*}
\subsection{Comparison to State-of-the-Art Methods}
We evaluate our Telescopic Adapters against a zero-shot baseline and multiple fine-tuning approaches across five medical datasets, as shown in Table~\ref{tab:medical_segmentation_results}. Fig.~\ref{fig:COMP} provides qualitative comparisons showing segmentation masks across representative samples from all datasets. The comparison includes E2E fine-tuning methods, LoRA-based adaptation, and existing adapter-based approaches, demonstrating significant performance and efficiency improvements.

CLIPSeg \cite{clipseg} (150M parameters) serves as our baseline VLSM, extending CLIP's vision-language capabilities to segmentation tasks. Its zero-shot performance, however, is substantially low (e.g., 53.42\% DSC on Kvasir-SEG), establishing a baseline that highlights the need for adaptation. While E2E fine-tuned CLIPSeg achieves competitive performance across datasets (87.69\% DSC on Kvasir-SEG, 91.88\% DSC on ISIC-16), it requires updating the entire parameter space. CRIS \cite{cris} (147M parameters) employs a CNN-based CLIP encoder for region retrieval-based segmentation, achieving strong results particularly on BKAI (92.62\% DSC) and ClinicDB (93.63\% DSC). However, CRIS's convolutional encoder is incompatible with adapter insertion, necessitating E2E fine-tuning and limiting its efficiency for resource-constrained deployment.

Parameter-efficient approaches offer substantial computational savings while maintaining competitive performance. I-MedSAM \cite{imedsam} (1.6M parameters) combines frequency and LoRA adapters within the SAM framework. Despite its efficiency, I-MedSAM shows mixed results across datasets, achieving 88.05\% DSC on BKAI but only 61.52\% on BUSI. SAN \cite{san} (8.4M parameters) attaches a parallel ViT network to frozen CLIP encoders, modeling segmentation as region recognition with mask proposal and attention bias branches. However, SAN demonstrates poor performance, with particularly low scores on BKAI (66.26\% DSC) and BUSI (45.61\% DSC), suggesting limited effectiveness for medical domain adaptation.

Existing adapter methods for VLSMs include CLIPSeg Dense Adapter (DA-VLC, 3.2M) and Shallow Adapter (SA-VLC, 4.2M) variants \cite{vlsmadap}, which apply uniform adaptation across vision, text, and conditioning pathways without depth-aware scaling. CLIPSeg DA-VLC achieves consistent performance across datasets (89.06\% DSC on Kvasir-SEG, 92.02\% DSC on ISIC-16), while SA-VLC shows slightly lower performance (86.80\% DSC on Kvasir-SEG, 91.55\% DSC on ISIC-16). Both methods require significantly more parameters than our approach while achieving inferior results. As shown in Fig.~\ref{fig:COMP}, our Telescopic Adapters produce segmentation masks that closely align with ground truth across diverse medical imaging modalities, demonstrating superior qualitative performance compared to existing methods.

\section{Conclusion}
In this work, we introduced Telescopic Adapters, a parameter-efficient fine-tuning technique that adapts vision language segmentation models to medical imaging domains using only 0.4\% of the baseline model's parameters. Our approach demonstrates competitive performance to end-to-end fine-tuning while significantly outperforming existing PEFT methods across multiple medical datasets. 
Our comprehensive analysis reveals that strategic adapter placement and progressive dimension scaling enable efficient domain transfer with minimal computational overhead. The learned scale parameters empirically confirm that shallow layers require minimal changes to their learned embeddings, while deeper transformer layers benefit from increased adaptation capacity for task-specific semantic understanding.
 This paradigm opens pathways to continual learning and multi-task learning frameworks for VLSMs, where specialized adapters can be trained for new datasets or tasks while keeping the core architecture frozen to prevent forgetting. Such adaptability is particularly valuable for medical image segmentation, where datasets are often limited in size and diverse in imaging modalities.
We encourage researchers to build upon this work and explore the broader applications of telescopic adapters across various domains and computer vision tasks.
\vspace{-0.25em}



\clearpage


 
{\small
\bibliographystyle{ieee_fullname}
\bibliography{wacv2024_fm+wsss_v3}

@String(CVPR= {IEEE Conf. Comput. Vis. Pattern Recog.})

@String(ICCV= {Int. Conf. Comput. Vis.})

@String(ECCV= {Eur. Conf. Comput. Vis.})

@String(ICLR = {Int. Conf. Learn. Represent.})

@String(AAAI = {AAAI})

@String(CVPR  = {CVPR})

@String(ICCV  = {ICCV})

@String(ECCV  = {ECCV})

@String(ICLR  = {ICLR})

@article{CAD,
author = {El-Baz, Ayman and Beache, Garth M. and Gimel′farb, Georgy and Suzuki, Kenji and Okada, Kazunori and Elnakib, Ahmed and Soliman, Ahmed and Abdollahi, Behnoush},
title = {Computer-Aided Diagnosis Systems for Lung Cancer: Challenges and Methodologies},
journal = {International Journal of Biomedical Imaging},
volume = {2013},
number = {1},
pages = {942353},
doi = {https://doi.org/10.1155/2013/942353},
url = {https://onlinelibrary.wiley.com/doi/abs/10.1155/2013/942353},
eprint = {https://onlinelibrary.wiley.com/doi/pdf/10.1155/2013/942353},
year = {2013}
}

@article{defibation,
   title={Medical image segmentation using deep learning: A survey},
   volume={16},
   ISSN={1751-9667},
   url={http://dx.doi.org/10.1049/ipr2.12419},
   DOI={10.1049/ipr2.12419},
   number={5},
   journal={IET Image Processing},
   publisher={Institution of Engineering and Technology (IET)},
   author={Wang, Risheng and Lei, Tao and Cui, Ruixia and Zhang, Bingtao and Meng, Hongying and Nandi, Asoke K.},
   year={2022},
   month=jan, pages={1243–1267} }

@inproceedings{general,
  title={Versatile medical image segmentation learned from multi-source datasets via model self-disambiguation},
  author={Chen, Xiaoyang and Zheng, Hao and Li, Yuemeng and Ma, Yuncong and Ma, Liang and Li, Hongming and Fan, Yong},
  booktitle={Proceedings of the IEEE/CVF Conference on Computer Vision and Pattern Recognition},
  pages={11747--11756},
  year={2024}
}

@article{diag,
author = {Fauw, Jeffrey and Ledsam, Joseph and Romera-Paredes, Bernardino and Nikolov, Stanislav and Tomasev, Nenad and Blackwell, Sam and Askham, Harry and Glorot, Xavier and O’Donoghue, Brendan and Visentin, Daniel and Driessche, George and Lakshminarayanan, Balaji and Meyer, Clemens and Mackinder, Faith and Bouton, Simon and Ayoub, Kareem and Chopra, Reena and King, Dominic and Karthikesalingam, Alan and Ronneberger, Olaf},
year = {2018},
month = {09},
pages = {},
title = {Clinically applicable deep learning for diagnosis and referral in retinal disease},
volume = {24},
journal = {Nature Medicine},
doi = {10.1038/s41591-018-0107-6}
}

@article{treat3,
  title={FocusNetv2: Imbalanced large and small organ segmentation with adversarial shape constraint for head and neck CT images},
  author={Gao, Yunhe and Huang, Rui and Yang, Yiwei and Zhang, Jie and Shao, Kainan and Tao, Changjuan and Chen, Yuanyuan and Metaxas, Dimitris N and Li, Hongsheng and Chen, Ming},
  journal={Medical Image Analysis},
  volume={67},
  pages={101831},
  year={2021},
  publisher={Elsevier}
}

@article{moni,
author = {Fischl, Bruce and Salat, David and Busa, Evelina and Albert, Marilyn and Dieterich, Megan and Haselgrove, Christian and Kouwe, Andre and Killiany, Ron and Kennedy, David and Klaveness, Shuna and Montillo, Albert and Makris, Nikos and Rosen, Bruce and Dale, Anders},
year = {2002},
month = {02},
pages = {341-55},
title = {Whole Brain Segmentation: Automated Labeling of Neuroanatomical Structures in the Human Brain},
volume = {33},
journal = {Neuron},
doi = {10.1016/S0896-6273(02)00569}
}

@inproceedings{data_prob,
  title={Enabling data diversity: efficient automatic augmentation via regularized adversarial training},
  author={Gao, Yunhe and Tang, Zhiqiang and Zhou, Mu and Metaxas, Dimitris},
  booktitle={International conference on information processing in medical imaging},
  pages={85--97},
  year={2021},
  organization={Springer}
}

@inproceedings{good,
  title={SEMI-Supervised Medical Image Segmentation via Dual Networks},
  author={Lu, Yunyao and Wu, Yihang and Kateb, Reem and Chaddad, Ahmad},
  booktitle={2025 IEEE 22nd International Symposium on Biomedical Imaging (ISBI)},
  pages={1--5},
  year={2025},
  organization={IEEE}
}

@inproceedings{suc5,
  title={Dodnet: Learning to segment multi-organ and tumors from multiple partially labeled datasets},
  author={Zhang, Jianpeng and Xie, Yutong and Xia, Yong and Shen, Chunhua},
  booktitle={Proceedings of the IEEE/CVF conference on computer vision and pattern recognition},
  pages={1195--1204},
  year={2021}
}

@InProceedings{suc6,
        author = { Zhong, Yuan and Tang, Chenhui and Yang, Yumeng and Qi, Ruoxi and Zhou, Kang and Gong, Yuqi and Heng, Pheng-Ann and Hsiao, Janet H. and Dou, Qi},
        title = { { Weakly-supervised Medical Image Segmentation with Gaze Annotations } },
        booktitle = {proceedings of Medical Image Computing and Computer Assisted Intervention -- MICCAI 2024},
        year = {2024},
        publisher = {Springer Nature Switzerland},
        volume = {LNCS 15003},
        month = {October},
        page = {530 -- 540}
}

@article{finetune3,
  title={Clip in medical imaging: A comprehensive survey},
  author={Zhao, Zihao and Liu, Yuxiao and Wu, Han and Wang, Mei and Li, Yonghao and Wang, Sheng and Teng, Lin and Liu, Disheng and Cui, Zhiming and Wang, Qian and others},
  journal={arXiv preprint arXiv:2312.07353},
  year={2023}
}

@inproceedings{cris,
  title={Cris: Clip-driven referring image segmentation},
  author={Wang, Zhaoqing and Lu, Yu and Li, Qiang and Tao, Xunqiang and Guo, Yandong and Gong, Mingming and Liu, Tongliang},
  booktitle={Proceedings of the IEEE/CVF conference on computer vision and pattern recognition},
  pages={11686--11695},
  year={2022}
}

@inproceedings{open4,
  title={Zegclip: Towards adapting clip for zero-shot semantic segmentation},
  author={Zhou, Ziqin and Lei, Yinjie and Zhang, Bowen and Liu, Lingqiao and Liu, Yifan},
  booktitle={Proceedings of the IEEE/CVF conference on computer vision and pattern recognition},
  pages={11175--11185},
  year={2023}
}

@inproceedings{finetunig,
  title={On pre-trained image features and synthetic images for deep learning},
  author={Hinterstoisser, Stefan and Lepetit, Vincent and Wohlhart, Paul and Konolige, Kurt},
  booktitle={Proceedings of the European Conference on Computer Vision (ECCV) Workshops},
  pages={0--0},
  year={2018}
}

@inproceedings{vnlm3,
  title={Text-guided cross-position attention for segmentation: Case of medical image},
  author={Lee, Go-Eun and Kim, Seon Ho and Cho, Jungchan and Choi, Sang Tae and Choi, Sang-Il},
  booktitle={International Conference on Medical Image Computing and Computer-Assisted Intervention},
  pages={537--546},
  year={2023},
  organization={Springer}
}

@inproceedings{clip,
  title={Learning transferable visual models from natural language supervision},
  author={Radford, Alec and Kim, Jong Wook and Hallacy, Chris and Ramesh, Aditya and Goh, Gabriel and Agarwal, Sandhini and Sastry, Girish and Askell, Amanda and Mishkin, Pamela and Clark, Jack and others},
  booktitle={International conference on machine learning},
  pages={8748--8763},
  year={2021},
  organization={PmLR}
}

@inproceedings{clipseg,
  title={Image segmentation using text and image prompts},
  author={L{\"u}ddecke, Timo and Ecker, Alexander},
  booktitle={Proceedings of the IEEE/CVF conference on computer vision and pattern recognition},
  pages={7086--7096},
  year={2022}
}

@article{vlsm2,
  title={VLSM-Net: A fusion architecture for CT image segmentation},
  author={Gao, Yachun and Guo, Jia and Fu, Chuanji and Wang, Yan and Cai, Shimin},
  journal={Applied Sciences},
  volume={13},
  number={7},
  pages={4384},
  year={2023},
  publisher={MDPI}
}

@article{vlsm3,
  title={Exploring transfer learning in medical image segmentation using vision-language models},
  author={Poudel, Kanchan and Dhakal, Manish and Bhandari, Prasiddha and Adhikari, Rabin and Thapaliya, Safal and Khanal, Bishesh},
  journal={arXiv preprint arXiv:2308.07706},
  year={2023}
}

@inproceedings{yu,
  title={Zero-shot referring image segmentation with global-local context features},
  author={Yu, Seonghoon and Seo, Paul Hongsuck and Son, Jeany},
  booktitle={Proceedings of the IEEE/CVF conference on computer vision and pattern recognition},
  pages={19456--19465},
  year={2023}
}

@article{lora,
  title={Lora: Low-rank adaptation of large language models.},
  author={Hu, Edward J and Shen, Yelong and Wallis, Phillip and Allen-Zhu, Zeyuan and Li, Yuanzhi and Wang, Shean and Wang, Lu and Chen, Weizhu and others},
  journal={ICLR},
  volume={1},
  number={2},
  pages={3},
  year={2022}
}

@inproceedings{adaptoer,
  title={Parameter-efficient transfer learning for NLP},
  author={Houlsby, Neil and Giurgiu, Andrei and Jastrzebski, Stanislaw and Morrone, Bruna and De Laroussilhe, Quentin and Gesmundo, Andrea and Attariyan, Mona and Gelly, Sylvain},
  booktitle={International conference on machine learning},
  pages={2790--2799},
  year={2019},
  organization={PMLR}
}

@article{ca,
  title={Clip-adapter: Better vision-language models with feature adapters},
  author={Gao, Peng and Geng, Shijie and Zhang, Renrui and Ma, Teli and Fang, Rongyao and Zhang, Yongfeng and Li, Hongsheng and Qiao, Yu},
  journal={International Journal of Computer Vision},
  volume={132},
  number={2},
  pages={581--595},
  year={2024},
  publisher={Springer}
}

@inproceedings{vlsmadap,
  title={Vlsm-adapter: Finetuning vision-language segmentation efficiently with lightweight blocks},
  author={Dhakal, Manish and Adhikari, Rabin and Thapaliya, Safal and Khanal, Bishesh},
  booktitle={International Conference on Medical Image Computing and Computer-Assisted Intervention},
  pages={712--722},
  year={2024},
  organization={Springer}
}

@article{ex_adap2,
  title={An efficient segment anything model for the segmentation of medical images},
  author={Dong, Guanliang and Wang, Zhangquan and Chen, Yourong and Sun, Yuliang and Song, Hongbo and Liu, Liyuan and Cui, Haidong},
  journal={Scientific Reports},
  volume={14},
  number={1},
  pages={19425},
  year={2024},
  publisher={Nature Publishing Group UK London}
}

@inproceedings{kvasir,
  title={Kvasir-seg: A segmented polyp dataset},
  author={Jha, Debesh and Smedsrud, Pia H and Riegler, Michael A and Halvorsen, P{\aa}l and De Lange, Thomas and Johansen, Dag and Johansen, H{\aa}vard D},
  booktitle={International conference on multimedia modeling},
  pages={451--462},
  year={2019},
  organization={Springer}
}

@article{clinicbd,
  title={WM-DOVA maps for accurate polyp highlighting in colonoscopy: Validation vs. saliency maps from physicians},
  author={Bernal, Jorge and S{\'a}nchez, F Javier and Fern{\'a}ndez-Esparrach, Gloria and Gil, Debora and Rodr{\'\i}guez, Cristina and Vilari{\~n}o, Fernando},
  journal={Computerized medical imaging and graphics},
  volume={43},
  pages={99--111},
  year={2015},
  publisher={Elsevier}
}

@inproceedings{bakai,
  title={Neounet: Towards accurate colon polyp segmentation and neoplasm detection},
  author={Ngoc Lan, Phan and An, Nguyen Sy and Hang, Dao Viet and Long, Dao Van and Trung, Tran Quang and Thuy, Nguyen Thi and Sang, Dinh Viet},
  booktitle={Advances in visual computing: 16th international symposium, ISVC 2021, virtual event, October 4-6, 2021, proceedings, part II},
  pages={15--28},
  year={2021},
  organization={Springer}
}

@article{isic16,
  title={Skin lesion analysis toward melanoma detection: A challenge at the international symposium on biomedical imaging (ISBI) 2016, hosted by the international skin imaging collaboration (ISIC)},
  author={Gutman, David and Codella, Noel CF and Celebi, Emre and Helba, Brian and Marchetti, Michael and Mishra, Nabin and Halpern, Allan},
  journal={arXiv preprint arXiv:1605.01397},
  year={2016}
}

@article{busi,
author = {Al-Dhabyani, Walid and Gomaa, Mohammed and Khaled, H.M. and Fahmy, Aly},
year = {2019},
month = {11},
pages = {104863},
title = {Dataset of Breast Ultrasound Images},
volume = {28},
journal = {Data in Brief},
doi = {10.1016/j.dib.2019.104863}
}

@article{adam,
  title={Decoupled weight decay regularization},
  author={Loshchilov, Ilya and Hutter, Frank},
  journal={arXiv preprint arXiv:1711.05101},
  year={2017}
}

@ARTICLE{SILU,
  author={Shah, Vivswan and Youngblood, Nathan},
  journal={IEEE Journal of Selected Topics in Quantum Electronics}, 
  title={Leveraging Continuously Differentiable Activation for Learning in Analog and Quantized Noisy Environments}, 
  year={2025},
  volume={31},
  number={3: AI/ML Integrated Opto-electronics},
  pages={1-9},
  doi={10.1109/JSTQE.2025.3534636}}

@article{SILU2,
  title={Deep network approximation: Beyond relu to diverse activation functions},
  author={Zhang, Shijun and Lu, Jianfeng and Zhao, Hongkai},
  journal={Journal of Machine Learning Research},
  volume={25},
  number={35},
  pages={1--39},
  year={2024}
}

@inproceedings{dim,
  title={Less is more: Pay less attention in vision transformers},
  author={Pan, Zizheng and Zhuang, Bohan and He, Haoyu and Liu, Jing and Cai, Jianfei},
  booktitle={Proceedings of the AAAI conference on artificial intelligence},
  volume={36},
  number={2},
  pages={2035--2043},
  year={2022}
}

@article{59, title={Mutual-Modality Adversarial Attack with Semantic Perturbation}, volume={38}, url={https://ojs.aaai.org/index.php/AAAI/article/view/28488}, DOI={10.1609/aaai.v38i7.28488}, number={7}, journal={Proceedings of the AAAI Conference on Artificial Intelligence}, author={Ye, Jingwen and Yu, Ruonan and Liu, Songhua and Wang, Xinchao}, year={2024}, month={Mar.}, pages={6657-6665} }

@inproceedings{60,
  title={Task Residual for Tuning Vision-Language Models},
  author={Yu, Tao and Lu, Zhihe and Jin, Xin and Chen, Zhibo and Wang, Xinchao},
  booktitle={Proceedings of the IEEE/CVF Conference on Computer Vision and Pattern Recognition},
  pages={10899--10909},
  year={2023}
}

@article{16,
  title={Multimodal neurons in artificial neural networks},
  author={Goh, Gabriel and Cammarata, Nick and Voss, Chelsea and Carter, Shan and Petrov, Michael and Schubert, Ludwig and Radford, Alec and Olah, Chris},
  journal={Distill},
  volume={6},
  number={3},
  pages={e30},
  year={2021}
}

@software{openclip,
  author       = {Ilharco, Gabriel and
                  Wortsman, Mitchell and
                  Wightman, Ross and
                  Gordon, Cade and
                  Carlini, Nicholas and
                  Taori, Rohan and
                  Dave, Achal and
                  Shankar, Vaishaal and
                  Namkoong, Hongseok and
                  Miller, John and
                  Hajishirzi, Hannaneh and
                  Farhadi, Ali and
                  Schmidt, Ludwig},
  title        = {OpenCLIP},
  month        = jul,
  year         = 2021,
  note         = {If you use this software, please cite it as below.},
  publisher    = {Zenodo},
  version      = {0.1},
  doi          = {10.5281/zenodo.5143773},
  url          = {https://doi.org/10.5281/zenodo.5143773}
}

@inproceedings{
18,
title={Open-vocabulary Object Detection via Vision and Language Knowledge Distillation},
author={Xiuye Gu and Tsung-Yi Lin and Weicheng Kuo and Yin Cui},
booktitle={International Conference on Learning Representations},
year={2022},
url={https://openreview.net/forum?id=lL3lnMbR4WU}
}

@inproceedings{14,
    title = "{P}ub{M}ed{CLIP}: How Much Does {CLIP} Benefit Visual Question Answering in the Medical Domain?",
    author = "Eslami, Sedigheh  and
      Meinel, Christoph  and
      de Melo, Gerard",
    editor = "Vlachos, Andreas  and
      Augenstein, Isabelle",
    booktitle = "Findings of the Association for Computational Linguistics: EACL 2023",
    month = may,
    year = "2023",
    address = "Dubrovnik, Croatia",
    publisher = "Association for Computational Linguistics",
    url = "https://aclanthology.org/2023.findings-eacl.88/",
    doi = "10.18653/v1/2023.findings-eacl.88",
    pages = "1181--1193",
}

@inproceedings{48,
    title = "{M}ed{CLIP}: Contrastive Learning from Unpaired Medical Images and Text",
    author = "Wang, Zifeng  and
      Wu, Zhenbang  and
      Agarwal, Dinesh  and
      Sun, Jimeng",
    editor = "Goldberg, Yoav  and
      Kozareva, Zornitsa  and
      Zhang, Yue",
    booktitle = "Proceedings of the 2022 Conference on Empirical Methods in Natural Language Processing",
    month = dec,
    year = "2022",
    address = "Abu Dhabi, United Arab Emirates",
    publisher = "Association for Computational Linguistics",
    url = "https://aclanthology.org/2022.emnlp-main.256/",
    doi = "10.18653/v1/2022.emnlp-main.256",
    pages = "3876--3887"
}

@article{biomed,
  title={A Multimodal Biomedical Foundation Model Trained from Fifteen Million Image–Text Pairs},
  author={Sheng Zhang and Yanbo Xu and Naoto Usuyama and Hanwen Xu and Jaspreet Bagga and Robert Tinn and Sam Preston and Rajesh Rao and Mu Wei and Naveen Valluri and Cliff Wong and Andrea Tupini and Yu Wang and Matt Mazzola and Swadheen Shukla and Lars Liden and Jianfeng Gao and Angela Crabtree and Brian Piening and Carlo Bifulco and Matthew P. Lungren and Tristan Naumann and Sheng Wang and Hoifung Poon},
  journal={NEJM AI},
  year={2024},
  volume={2},
  number={1},
  doi={10.1056/AIoa2400640},
  url={https://ai.nejm.org/doi/full/10.1056/AIoa2400640}
}

@inproceedings{medclip,
  title={Medclip: Contrastive learning from unpaired medical images and text},
  author={Wang, Zifeng and Wu, Zhenbang and Agarwal, Dinesh and Sun, Jimeng},
  booktitle={Proceedings of the Conference on Empirical Methods in Natural Language Processing. Conference on Empirical Methods in Natural Language Processing},
  volume={2022},
  pages={3876},
  year={2022}
}

@article{82,
  author       = {Youshan Zhang},
  title        = {A Survey of Unsupervised Domain Adaptation for Visual Recognition},
  journal      = {CoRR},
  volume       = {abs/2112.06745},
  year         = {2021},
  url          = {https://arxiv.org/abs/2112.06745},
  eprinttype    = {arXiv},
  eprint       = {2112.06745},
  bibsource    = {dblp computer science bibliography, https://dblp.org}
}

@INPROCEEDINGS{43,
  author={Zou, Yang and Yu, Zhiding and Liu, Xiaofeng and Kumar, B. V. K. Vijaya and Wang, Jinsong},
  booktitle={2019 IEEE/CVF International Conference on Computer Vision (ICCV)}, 
  title={Confidence Regularized Self-Training}, 
  year={2019},
  volume={},
  number={},
  pages={5981-5990},
  doi={10.1109/ICCV.2019.00608}}

@article{37,
  title={Domain stylization: A strong, simple baseline for synthetic to real image domain adaptation},
  author={Dundar, Aysegul and Liu, Ming-Yu and Wang, Ting-Chun and Zedlewski, John and Kautz, Jan},
  journal={arXiv preprint arXiv:1807.09384},
  year={2018}
}

@InProceedings{62,
    author    = {VS, Vibashan and Oza, Poojan and Patel, Vishal M.},
    title     = {Towards Online Domain Adaptive Object Detection},
    booktitle = {Proceedings of the IEEE/CVF Winter Conference on Applications of Computer Vision (WACV)},
    month     = {January},
    year      = {2023},
    pages     = {478-488}
}

@inproceedings{3,
      title = {To Adapt or Not to Adapt? Real-Time Adaptation for Semantic Segmentation},
      author = {Botet Colomer, Marc and 
                Dovesi, Pier Luigi and 
                Panagiotakopoulos, Theodoros and 
                Carvalho, Joao Frederico and 
                H{\"a}renstam-Nielsen, Linus and 
                Azizpour, Hossein and 
                Kjellstr{\"o}m, Hedvig and 
                Cremers, Daniel and
                Poggi, Matteo},
      booktitle = {IEEE International Conference on Computer Vision},
      note = {ICCV},
      year = {2023}
}

@inproceedings{611,
author = {Panagiotakopoulos, Theodoros and Dovesi, Pier Luigi and H\"{a}renstam-Nielsen, Linus and Poggi, Matteo},
title = {Online Domain Adaptation for;Semantic Segmentation in;Ever-Changing Conditions},
year = {2022},
isbn = {978-3-031-19829-8},
publisher = {Springer-Verlag},
address = {Berlin, Heidelberg},
url = {https://doi.org/10.1007/978-3-031-19830-4_8},
doi = {10.1007/978-3-031-19830-4_8},
booktitle = {Computer Vision – ECCV 2022: 17th European Conference, Tel Aviv, Israel, October 23–27, 2022, Proceedings, Part XXXIV},
pages = {128–146},
numpages = {19},
location = {Tel Aviv, Israel}
}

@inproceedings{bitfit,
    title = "{B}it{F}it: Simple Parameter-efficient Fine-tuning for Transformer-based Masked Language-models",
    author = "Ben Zaken, Elad  and
      Goldberg, Yoav  and
      Ravfogel, Shauli",
    editor = "Muresan, Smaranda  and
      Nakov, Preslav  and
      Villavicencio, Aline",
    booktitle = "Proceedings of the 60th Annual Meeting of the Association for Computational Linguistics (Volume 2: Short Papers)",
    month = may,
    year = "2022",
    address = "Dublin, Ireland",
    publisher = "Association for Computational Linguistics",
    url = "https://aclanthology.org/2022.acl-short.1/",
    doi = "10.18653/v1/2022.acl-short.1",
    pages = "1--9",
   
}

@article{prune2,
  title={Nxmtransformer: Semi-structured sparsification for natural language understanding via admm},
  author={Holmes, Connor and Zhang, Minjia and He, Yuxiong and Wu, Bo},
  journal={Advances in neural information processing systems},
  volume={34},
  pages={1818--1830},
  year={2021}
}

@inproceedings{fusion,
    title = "{A}dapter{F}usion: Non-Destructive Task Composition for Transfer Learning",
    author = {Pfeiffer, Jonas  and
      Kamath, Aishwarya  and
      R{\"u}ckl{\'e}, Andreas  and
      Cho, Kyunghyun  and
      Gurevych, Iryna},
    editor = "Merlo, Paola  and
      Tiedemann, Jorg  and
      Tsarfaty, Reut",
    booktitle = "Proceedings of the 16th Conference of the European Chapter of the Association for Computational Linguistics: Main Volume",
    month = apr,
    year = "2021",
    address = "Online",
    publisher = "Association for Computational Linguistics",
    url = "https://aclanthology.org/2021.eacl-main.39/",
    doi = "10.18653/v1/2021.eacl-main.39",
    pages = "487--503",
}

@inproceedings{prompt1,
author = {Jia, Menglin and Tang, Luming and Chen, Bor-Chun and Cardie, Claire and Belongie, Serge and Hariharan, Bharath and Lim, Ser-Nam},
title = {Visual Prompt Tuning},
year = {2022},
isbn = {978-3-031-19826-7},
publisher = {Springer-Verlag},
address = {Berlin, Heidelberg},
url = {https://doi.org/10.1007/978-3-031-19827-4_41},
doi = {10.1007/978-3-031-19827-4_41},

booktitle = {Computer Vision – ECCV 2022: 17th European Conference, Tel Aviv, Israel, October 23–27, 2022, Proceedings, Part XXXIII},
pages = {709–727},
numpages = {19},
location = {Tel Aviv, Israel}
}

@inproceedings{fs1,
  author    = {Manli, Shu and Weili, Nie and De-An, Huang and Zhiding, Yu and Tom, Goldstein and Anima, Anandkumar and Chaowei, Xiao},
  title     = {Test-Time Prompt Tuning for Zero-shot Generalization in Vision-Language Models},
  booktitle = {NeurIPS},
  year      = {2022},
}

@unknown{333,
author = {Lialin, Vladislav and Deshpande, Vijeta and Rumshisky, Anna},
year = {2023},
month = {03},
pages = {},
title = {Scaling Down to Scale Up: A Guide to Parameter-Efficient Fine-Tuning},
doi = {10.48550/arXiv.2303.15647}
}

@inproceedings{lora2,
title={Adaptive Budget Allocation for Parameter-Efficient Fine-Tuning },
author={Qingru Zhang and Minshuo Chen and Alexander Bukharin and Pengcheng He and Yu Cheng and Weizhu Chen and Tuo Zhao},
booktitle={The Eleventh International Conference on Learning Representations },
year={2023},
url={https://openreview.net/forum?id=lq62uWRJjiY}
}

@article{perform1,
title = {Corrigendum to “Hydra: Multi-head Low-rank Adaptation for Parameter Efficient Fine-tuning” [Neural Networks Volume 178, October (2024), 1-11/106414]]},
journal = {Neural Networks},
volume = {181},
pages = {106878},
year = {2025},
issn = {0893-6080},
doi = {https://doi.org/10.1016/j.neunet.2024.106878},
url = {https://www.sciencedirect.com/science/article/pii/S0893608024008074},
author = {Sanghyeon Kim and Hyunmo Yang and Younghyun Kim and Youngjoon Hong and Eunbyung Park}
}

@misc{
perform2,
title={One-for-All: Generalized Lo{RA} for Parameter-Efficient Fine-tuning},
author={Arnav Chavan and Zhuang Liu and Deepak Gupta and Eric Xing and Zhiqiang Shen},
year={2024},
url={https://openreview.net/forum?id=K7KQkiHanD}
}

@InProceedings{imedsam,
author="Wei, Xiaobao
and Cao, Jiajun
and Jin, Yizhu
and Lu, Ming
and Wang, Guangyu
and Zhang, Shanghang",
editor="Leonardis, Ale{\v{s}}
and Ricci, Elisa
and Roth, Stefan
and Russakovsky, Olga
and Sattler, Torsten
and Varol, G{\"u}l",
title="I-MedSAM: Implicit Medical Image Segmentation with Segment Anything",
booktitle="Computer Vision -- ECCV 2024",
year="2025",
publisher="Springer Nature Switzerland",
address="Cham",
pages="90--107",
isbn="978-3-031-72684-2"
}

@InProceedings{san,
    author    = {Xu, Mengde and Zhang, Zheng and Wei, Fangyun and Hu, Han and Bai, Xiang},
    title     = {Side Adapter Network for Open-Vocabulary Semantic Segmentation},
    booktitle = {Proceedings of the IEEE/CVF Conference on Computer Vision and Pattern Recognition (CVPR)},
    month     = {June},
    year      = {2023},
    pages     = {2945-2954}
}
}

\end{document}